\pdfoutput=1

\documentclass[11pt]{article}
\usepackage{CJKutf8}
\usepackage{amsmath}
\usepackage[review]{ACL2023}

\usepackage{times}
\usepackage{latexsym}
\usepackage[T1]{fontenc}

\usepackage[utf8]{inputenc}

\usepackage{microtype}

\usepackage{inconsolata}

\usepackage{graphicx}
\usepackage{booktabs}
\usepackage{multirow}

\title{Assisting Language Learners: Automated Trans-Lingual Definition Generation via Contrastive Prompt Learning}

\renewcommand\footnotemark{}
\author{\textbf{Hengyuan Zhang\textsuperscript{1}, Dawei Li\textsuperscript{2}, Yanran Li\textsuperscript{3\dag}}\thanks{\textsuperscript{\dag} Corresponding author}, Chenming Shang\textsuperscript{1}, Chufan Shi\textsuperscript{1}, Yong Jiang\textsuperscript{1} \\
  \textsuperscript{1}Shenzhen International Graduate School, Tsinghua University \\
  \textsuperscript{2}Halicioğlu Data Science Institute, University of California, San Diego \\
  \textsuperscript{3}Independent Researcher \\
  \texttt{zhang-hy22@mails.tsinghua.edu.cn}\\
}

\begin{document}
\maketitle
\begin{abstract}
The standard definition generation task requires to automatically produce mono-lingual definitions (e.g., English definitions for English words), but ignores that the generated definitions may also consist of unfamiliar words for language learners. In this work, we propose a novel task of \textbf{T}rans-\textbf{L}ingual \textbf{D}efinition \textbf{G}eneration (TLDG), which aims to generate definitions in another language, i.e., the native speaker's language. Initially, we explore the unsupervised manner of this task and build up a simple implementation of fine-tuning the multi-lingual machine translation model. 
Then, we develop two novel methods, Prompt Combination and Contrastive Prompt Learning, for further enhancing the quality of the generation. Our methods are evaluated against the baseline Pipeline method in both rich- and low-resource settings, and we empirically establish its superiority in generating higher-quality trans-lingual definitions.
The ablation studies and further analysis are also conducted to provide more hints on this new task.
\end{abstract}

\section{Introduction}
A significant area of research within Intelligent Computer-Assisted Language Learning (ICALL) is devoted to supporting language learners in understanding words~\cite{ICALL2020-1,ICALL2020-2}. This research is primarily motivated by two main issues: (1) Language learners often struggle to accurately identify the meaning of words with multiple definitions, as the cognitive process of differentiating each meaning can be challenging~\cite{tyler2001reconsidering}; (2) On another note, lexicographers are responsible for manually updating predefined word-definition inventories for language learners, a process that may be time-consuming and not always able to keep up with the constantly evolving nature of language usage. To address these issues, researchers aim to benefit both language learners and lexicographers by automatically generating the definition for a given word based on its corresponding local context~\cite{ni-wang-2017-learning,gadetsky2018conditional,ishiwatari2019learning,bevilacqua2020generationary}. 

\begin{figure}[!t]
    \centering
    \includegraphics[scale=1.2]{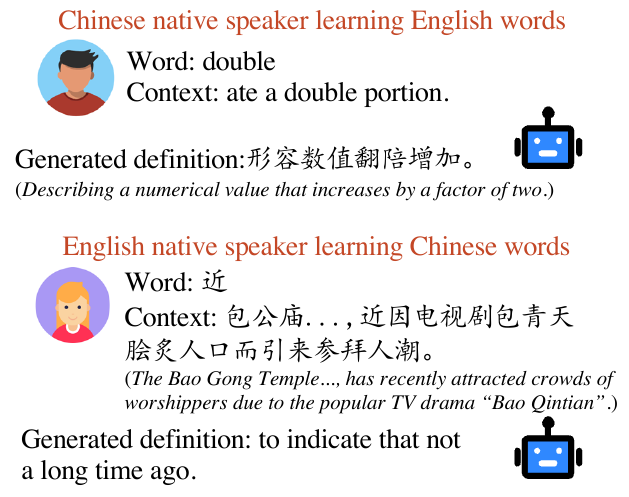}
    \caption{The application scenes of a Chinese native speaker learning English and English native speaker learning Chinese. We also build a Chrome extension (in Appendix~\ref{Chrome Extension Application Scene}) to better show the application scenes.}
    \vspace{-0.4cm}
    \label{application}
\end{figure}

Previous works on definition generation mainly focus on mono-lingual generation scenarios, primarily due to the availability of parallel training and evaluation data~\cite{yang2020incorporating,huang2021definition,zhang2022fine}. 
However, these works rarely notice a real-occurring problem that the generated definitions may also consist of unfamiliar words for language learners~\cite{zhang2011discussion}. In other words, it is more applicable to generate definitions in the native language of foreign learners. As depicted in Figure~\ref{application}, if a Chinese native speaker wants to know an English word's meaning, the definition in Chinese is easier to capture. 

To this end, we propose a novel task called \textbf{T}rans-\textbf{L}ingual \textbf{D}efinition \textbf{G}eneration (\textbf{TLDG}). The TLDG task is challenging because there are no trans-lingual parallel datasets, e.g., the word and context are in Chinese, and the definition is in English. Also, building trans-lingual parallel datasets is labor-consuming. 
To address this, we leverage the data resources of mono-lingual definition generation and utilize translation model to explore the trans-lingual definition generation task in an unsupervised manner. During preliminary experiments, we find two typical types of errors in the generated results. As shown in Table~\ref{error type}\footnote{In this paper, Zh-En means the input's word and context are in Chinese, and the expected generated definition is in English. Other language combinations are also similar.}, \emph{Ignore-task error} means the model only translates the input's context but neglects the definition generation task. \emph{Language-mix error} means words in different languages simultaneously appear in the generated definition.

To mitigate the problems, we develop two novel learning methods. For the Ignore-task error, we get inspired from task-oriented prompt learning~\cite{chung2022scaling,akyurek-etal-2022-measuring}, and design Prompt Combination method to force the models focus on generating trans-lingual definition rather than mere translation.
In addition, we propose Contrastive Prompt Learning method based on an contrastive loss~\cite{contrastive_loss,triplet-loss}, which separates language information from the task prompt and in turn acquires a better task prompt representation for definition generation. 
Due to the scarcity of definition generation data in numerous languages, we carry out extensive experiments in both rich- and low-resource situations. We demonstrate that the Contrastive Prompt Learning method is effective in addressing the two errors and capable of yielding higher-quality definitions when compared to the baselines in both scenarios.


\begin{table}[]\small
\centering
\setlength{\tabcolsep}{0.5mm}{
\begin{tabular}{@{}ccc@{}}
\hline
Context                                                                      & Generated Definition                                                           & Error Type                                                   \\ \hline
\begin{tabular}[c]{@{}l@{}}This food \ \underline{revitalized}\\ the patient.\end{tabular} 
    & \begin{tabular}[c]{@{}l@{}}\begin{CJK*}{UTF8}{gbsn}食物使病人恢复活 \end{CJK*} \\\begin{CJK*}{UTF8}{gbsn}力。\end{CJK*} (\textit{Food revitalizes} \\\textit{ patients}) \end{tabular}                 
    & \begin{tabular}[c]{@{}c@{}}Ignore-task\\ error\end{tabular}  \\ \hline
    \begin{tabular}[c]{@{}l@{}}\begin{CJK*}{UTF8}{gbsn}..., 各家各派对人性\end{CJK*}\\
    \begin{CJK*}{UTF8}{gbsn}的\underline{看法}极为不同。\end{CJK*}\\ (\textit{...,Each party has a }\\ \textit{very different view of }\\ \textit{human nature.}) \end{tabular}
    & \begin{tabular}[c]{@{}l@{}}\begin{CJK*}{UTF8}{gbsn}形容\end{CJK*} (\textit{Describe}) a \\person's opinion about \\something.\end{tabular}
    & \begin{tabular}[c]{@{}c@{}}Language-mix\\ error\end{tabular} \\ \hline
\end{tabular}}
\caption{Zh-En and En-Zh examples of the two error types in the unsupervised TLDG task. The target words are marked with \underline{underline} in context.}
\label{error type}
\end{table}

In general, our contributions are as follows:
\begin{itemize}
\item To better assist language learners, we propose the task of TLDG in an unsupervised manner and identify two typical errors.

\item We develop several methods to mitigate the problems and demonstrate the Contrastive Prompt Learning method yields promising performances in both rich- and low-resource scenarios.


\item We analyze the methods through ablated and case studies, and provide several hints on this newly introduced task. Also, we build a Chrome extension (in Appendix~\ref{Chrome Extension Application Scene}) to further show the application scene of our proposed task.
\end{itemize}
\section{Related Work}
\label{Related Work}
\subsection{Definition Generation}
\label{Definition Generation}
The task of definition generation is first proposed by \citet{noraset2017definition}, which aims to generate definitions from corresponding word embeddings.
Subsequent studies have investigated a broader range of application scenarios and model architectures for generating definitions.
To generate appropriate definitions for polysemies,~\citet{ni-wang-2017-learning} first introduce the context and input the context with the target word to a bi-encoder model.
Following them,~\citet{ishiwatari2019learning} develop a method that incorporates a gate mechanism in the decoding stage to integrate the information of the word and context.
There are also some works that try to model the semantic representation in a more detailed way. 
Specifically, they break down the meaning of the target word into several components and provide a fine-grained word representation for the generation stage~\cite{li2020explicit,reid2020vcdm}. 

Recently, some works adopt pre-trained encoder-decoder models in definition generation and achieved great success.
~\citet{huang2021definition} use a re-ranking strategy to obtain proper specific definitions.~\citet{zhang2022fine} regard word and definition as a semantic equivalence pair to do contrastive learning.
However, all the aforementioned works focus on improving the quality of the generated definitions, and the difficulty of understanding the definition itself for language learners has been ignored.

Although~\citet{kong2022multitasking} design a multi-task framework to generate definitions with more simple words, we argue that other factors like language grammar will still hinder language learners to understand the definition. To mitigate it, we propose a novel task of trans-lingual definition generation to generate definitions in the target language.

\subsection{Prompt Learning}
\label{Prompt Learning}
In recent years, numerous pre-trained models have been introduced, e.g., GPT~\cite{gpt}, BART~\cite{bart}. 
To adapt these models for different downstream tasks, prompt learning has been widely used. \citet{PET} manually design discrete template prompts to transform the downstream task into the text-infilling task, which is closer to the pre-trained paradigm.
Besides, in the conditional text generation field, both \citet{zhang2022persona} and \citet{xie2022psychology} regard attribute keywords as hard prompts and fuse them into the model to control the generation result. 
However, Manually designing hard prompts can be both tedious and challenging, later works suggest using the soft prompts that consist of multiple learnable embeddings for the downstream tasks~\cite{prefix-tuning,liu2021gpt,han2022ptr}. 

Furthermore, some works propose that rather than updating the entire PLM, it is more effective to fix its parameters and only update the soft prompts~\cite{prompt_tuning,qin2021learning}.
When using large PLMs as the backbone, this method can achieve comparable results to fine-tuning the entire model.
In the low-resource scenario, \citet{ppt} apply prompt initialization and use several tasks to obtain generalized prompts for different downstream tasks. \citet{zheng2021exploring} and \citet{zhang2021differentiable} use the prompt learning strategy to get different task-oriented prompts with corresponding task-specific objectives and achieve satisfactory results.

In this work, we use prompt learning to indicate the task and address the Ignore-task error. By developing a novel contrastive prompt learning loss, we finally achieve promising performances on both rich- and low-resource TLDG.

\begin{figure*}[!t]
    \centering
    \includegraphics[width=16cm]{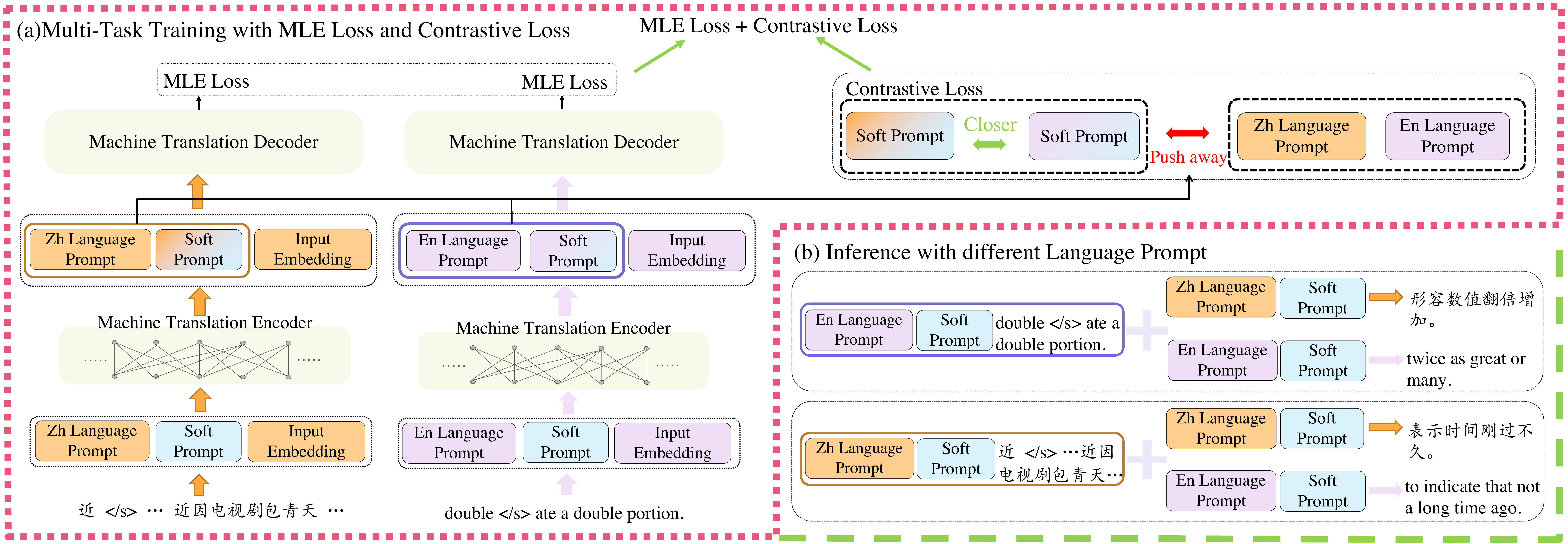}
    \caption{The area surrounded by the \textcolor[RGB]{244,67,54}{red} dotted line represents the training process and the \textcolor[RGB]{76,175,80}{green} dash line represents the inference process. In the training phase, (1) the task prompt will mix the language information from the language prompt (in blended color), and (2) the contrastive loss (upper right corner) is together applied with the MLE loss (upper left corner) to train the model jointly. At the inference stage, the language prompt could be set to any other languages used in the training stage for trans-lingual definition generation. Best viewed in color.}
    \label{overview}
\end{figure*}

\section{Method}
One straightforward approach to generating trans-lingual definitions is to develop a pipeline that initially produces mono-lingual definitions and then translates them into the desired language. This intuitive approach serves as one naive baseline, which we elaborate in the experiment section (Section~\ref{Baseline Pipeline Strategy}).

Besides, in this section, we introduce 3 methods to better fit our task: (1) a simple implementation of fine-tuning on multi-lingual translation model; (2) Prompt Combination method; and (3) Contrastive Prompt Learning method.
\label{Method}
\subsection{Task Formulation}
\label{Task Formulation}

The standard definition generation (DG) task is to generate the definition $D = \{d_0,...d_t\}$ for a given word or phrase $W = \{w_i,...,w_j\}$ and its corresponding context $C = \{w_0,...,w_k\}(0<i<j<k)$. 
Here, the context is a sentence containing the word. 
Note that standard DG is a mono-lingual task where the word, context, and definition are in the same language.

Distinguishedly, the task of trans-lingual definition generation (TLDG) is to generate trans-lingual definition $D_{l_{j}}$ in language $l_j$ for a given word $W_{l_{i}}$ and context $C_{l_{i}}$ in another language $l_i$. 
Since there does not exist TLDG example triplets $\{(W_{l_{i}}, C_{l_{i}}, D_{l_j})\}$, the only available resources are mono-lingual definition generation datasets. 
Hence, the TLDG task in this work can be regarded as a fully unsupervised task.

\subsection{Simple Implementation of Directly Fine-tuning Translation Model}
\label{Simple Implementation with Directly Fine-tuning Translation Model}
The newly introduced TLDG task aims to generate the trans-lingual definition without supervised parallel datasets. 
As neural machine translation (NMT) shows powerful performance in translation, as a preliminary attempt, we directly fine-tune multi-lingual NMT with existing mono-lingual DG datasets ($G$). Concretely, we concatenate language prompt (which is predefined in the multi-lingual NMT model to specify the source and target languages), target word, and context ($[L_{l_i}; W_{l_i}; C_{l_i}]$) as input $X_{l_i}$ to the encoder.
Similarly, we concatenate language prompt and definition ($[L_{l_i}; D_{l_i}]$) as ground-truth ${Y_{l_i}}$ to train the model, which can be formulated as:
\begin{equation}
\label{training trans-lingual task formulation}
P(Y_{l_i}|X_{l_i}) = \prod \limits_{t} p(y_t|y_{<t},X_{l_i};\theta)
\end{equation}
where $y_{t}$ is the $t$-th token of ${Y_{l_i}}$, $\theta$ is the
model's parameters to be tuned. 
To optimize, a cross-entropy loss is utilized to assess the difference between the distribution generated by the model and the ground-truth distribution, and the loss function is as follows:
\begin{equation}
\label{MLE Loss}
\mathcal{L}_{MLE} = - \sum_{(W_{l_i}, C_{l_i}, D_{l_i}) \in G_{l_i},\atop l_i \in L}  logP(Y_{l_i}|X_{l_i};\theta)
\end{equation}

By concatenating the corresponding language prompt, the model is able to infer trans-lingual definitions in any language previously seen in the training stage $(<W_{l_i}, C_{l_i}>\rightarrow D_{l_j}, l_i, l_j \in L)$.



\subsection{Prompt Combination}
\label{Prompt Combination}

Despite that fine-tuning the translation model seems plausible for trans-lingual definition generation, we find a plethora of Ignore-task cases in the generated definitions. 
We conjecture that the language prompt would still instinctively induce the translation model to perform the translation task, and thus leading to those Ignore-task errors.

To notify the model focus on the definition generation task, we add a specific task-oriented prompt after the language prompt. We adopt soft prompts for our task since they have been shown more flexible than hard prompts~\cite{p-tuning}. 
In the training stage, we insert the task prompt $T = \{t_1,t_2,..., t_n\}$ after the language prompt $L_{l_i}$ for both encoder and decoder inputs, where $n$ is the number of soft prompt tokens.

While this strategy mitigates the Ignore-task error in trans-lingual definition generation, we find adding the task-oriented soft prompt will lead to Language-mix errors. 
One possible explanation is that during the training stage, the task prompt is mixed up with the language information from the language prompt. 
During inference, such mixed task prompts will confuse the model to generate words in undesired languages.

\subsection{Contrastive Prompt Learning}
\label{Contrastive Prompt Learning}

To tackle this problem, we propose a Contrastive Prompt Learning method. This method aims to obtain a more informative and representative task prompt by decoupling the language information inside within it. The overview of the proposed method is illustrated in Figure~\ref{overview}, where we take Chinese and English as examples. 


In each batch, we randomly fetch training samples in two different languages ($l_i$ and $l_j$) and separate them into two groups.
After passing each group into the model, we extract the language prompt embedding $\mathbf{H}^{lp}_{l_i}$ and the task prompt embedding $\mathbf{H}^{tp}_{l_i}$ from each group's encoding $\mathbf{H}_{l_i}$ and $\mathbf{H}_{l_j}$ according to their positions:

\begin{equation}\label{embedding l1}
\mathbf{H}^{tp}_{l_i}, \mathbf{H}^{lp}_{l_i} = \textrm{Extract}(\mathbf{H}_{l_i})
\end{equation}

\begin{equation}\label{embedding l2}
\mathbf{H}^{tp}_{l_j}, \mathbf{H}^{lp}_{l_j} = \textrm{Extract}(\mathbf{H}_{l_j})
\end{equation}

Since the language prompt only has one token, we directly regard language prompt embedding as language prompt representation $\mathbf{h}^{lp}_{l_i}$. For multiple task prompt tokens, we apply the pooling function to $\mathbf{H}^{tp}_{l_i}$ and $\mathbf{H}^{tp}_{l_j}$ to get the task prompt representation $\mathbf{h}^{tp}_{l_i}$ and $\mathbf{h}^{tp}_{l_j}$. Without loss of generality, we implement attention-pooling, mean-pooling and max-pooling, and compare them in Section~\ref{Ablation Study}.

To build up contrastive loss, we regard task prompt representation in different languages as positive pairs~$(\mathbf{h}^{tp}_{l_i}$,  $\mathbf{h}^{tp}_{l_j})$, task prompt representation and different language prompt representation as negative pairs~$\{(\mathbf{h}^{tp}_{l_i}$, $\mathbf{h}^{lp}_{l_j}),\ l_i, l_j \in L\}$.
By doing so, the language information in $\mathbf{h}^{tp}_{l_i}$ and $\mathbf{h}^{tp}_{l_j}$ can be effectively eliminated. 
Mathematically, the contrastive loss is formulated as:

\begin{equation}
\label{contrastive loss}
\mathcal{L}_C = \max(d_p - d_n + \sigma, 0) / \tau
\end{equation}

\begin{equation}
\label{contrastive loss2}
\begin{split}
d_p &= \|\mathbf{h}^{tp}_{l_i}-\mathbf{h}^{tp}_{l_j}\|  \\
d_n &= \sum_{a \in \{i, j\}}{\frac{1}{2}\|\mathbf{h}^{tp}_{l_i}-\mathbf{h}^{lp}_{l_a}\|}
\end{split}
\end{equation}
where $d_p$ is the distance of positive pair, $d_n$ is the average distance of negative pairs, $\sigma$ is the margin and $\tau$ is the temperature to scale the contrastive loss. 

As Figure~\ref{overview} depicts, the proposed contrastive loss is combined with MLE loss to train the model:

\begin{equation}
\label{eq:Final_Loss}
\mathcal{L}_{Final} = \lambda*\mathcal{L_C} + (1-\lambda)*\mathcal{L}_{MLE}
\end{equation}
where $\lambda$ is a hyper-parameter to balance the two losses. In this way, our method is able to (1) separate the language information from the task prompt based on the novel contrastive loss, and (2) obtain a more oriented and pure task prompt representation for generating trans-lingual definition.

\section{Experiments}
In this section, we conduct extensive experiments and analyze the proposed methods carefully.
\subsection{Datasets}
Considering that many languages do not have sufficient definition generation data, we validate the proposed method in both rich- and low-resource scenarios. Note that all the datasets we use to train models are the mono-lingual definition generation datasets, which means the source language and target language are the same. 
\paragraph{Rich-resource}
In the rich-resource scenario, we train and evaluate our models using English and Chinese definition generation datasets. For English, we use the Oxford dataset, collected using Oxford APIs of Oxford Dictionary\footnote{\url{https://developer.oxforddictionaries.com}} by~\citet{gadetsky2018conditional}.
We follow~\citet{ishiwatari2019learning} to split them into training, validation, and test sets.

For Chinese, we follow~\citet{kong-etal-2022-multitasking} to use Chinese-WordNet (CWN)~\cite{huang2010chinese} and split them into training, validation, and test sets. 
It is a semantic lexicon aiming to provide a knowledge base of sense\footnote{\url{https://lope.linguistics.ntu.edu.tw/cwn2/}}.
The statistics of these two datasets are shown in Appendix~\ref{Rich-resource Detailed Dataset Setting}.
In the inference stage, we conduct En-Zh, Zh-En trans-lingual definition generation.

\paragraph{Low-resource}
In the low-resource scenario, we set the training data size to 256, validation data size to 200, and following~\citet{few-shot-1,few-shot-2} to use the validation set as test set. 

In specific, we build few-shot mono-lingual training datasets in English, Chinese, and France. For English and Chinese, we randomly choose samples from Oxford and CWN. For France, as there doesn't exist any public France definition generation dataset, we follow~\citet{vcdm} to collect data from Lerobert Dictionary\footnote{\url{https://dictionnaire.lerobert.com}}.
In the inference stage, we conduct trans-lingual definition generation with 6 settings, i.e., En-Zh, Zh-En, En-Fr, Fr-En, Zh-Fr, and Fr-Zh.

\subsection{Experimental Settings}
\label{Experimental Settings}
In this work, we utilize two multi-lingual NMT models, namely mBART-many-to-many\footnote{\url{https://huggingface.co/facebook/mbart-large-50-many-to-many-mmt}} (a model that fine-tuned on mBART~\cite{liu2020multilingual} with downstream machine translation tasks) and M2M\footnote{\url{https://huggingface.co/facebook/m2m100_418M}} (a model that directly trained on massive multi-lingual translation tasks from scratch), to implement our ideas. For convenience, we use mBART-T to represent mBART-many-to-many in this paper.

For all experiments, we set the batch size to 16 and use Adam optimizer to update parameters. We train all of our models on a V100 GPU. Following~\citet{prompt_tuning}, we adopt 100 tunable soft prompt tokens. For the Contrastive Prompt Learning method, we set the temperature as 0.16 to scale the contrastive loss. The best performances in Section~\ref{Main Results} adopt the attention-pooling function.
\begin{table*}[!h]\small
\fontsize{10pt}{8pt}\selectfont
\centering
\setlength{\tabcolsep}{8.0mm}{
\begin{tabular}{@{}cccccccc@{}}
\toprule
\multirow{2}{*}{Model}                  & \multirow{2}{*}{Method}             & \multicolumn{2}{c}{Semantic Sim} \\ \cmidrule(l){3-4} 

\multirow{4}{*}{mBART + M2M}   &  & En-Zh              & Zh-En                          \\ \midrule
                        & Pipeline & 47.58 & 52.24 \\ \midrule
\multirow{3}{*}{M2M}   & w/ Directly Fine-tuning          & 45.69              & 51.68              \\
                       & w/ Prompt Combination      & 47.56              & 52.63              \\
                       & w/ Contrastive Prompt Learning      & \textbf{49.42}     & \textbf{55.19}     \\ \midrule
\multirow{3}{*}{mBART-T} & w/ Directly Fine-tuning           & 43.32              & 51.16              \\
                       & w/ Prompt Combination        & 45.13              & 51.85              \\
                       & w/ Contrastive Prompt Learning     & 47.79             & 53.92              \\ \midrule 
\end{tabular}
}
\caption{Automatic evaluation results on the rich-resource test dataset. The best results are in \textbf{bold}.}
\vspace{-5mm}
\label{Automatic-results}
\end{table*}
\paragraph{Compared Methods}
\label{Baseline Pipeline Strategy} 
We compare with four methods: (1) A naive \textbf{Pipeline} method; (2) \textbf{Directly Fine-tuning} (Section~\ref{Simple Implementation with Directly Fine-tuning Translation Model}); (3) \textbf{Prompt Combination} (Section~\ref{Prompt Combination}); (4) \textbf{Contrastive Prompt} (Section~\ref{Contrastive Prompt Learning}). Specifically, the Pipeline method consists of generation and translation procedures. We begin with fine-tuning the pre-trained mBART~\cite{liu2020multilingual} model with mono-lingual datasets to generate mono-lingual definitions rather than trans-lingual definitions. Subsequently, we utilize the M2M model to translate the generated definitions into the target language. 
\paragraph{Rich-resource}
In the rich-resource scenario, we fine-tune all the parameters (including soft prompt tokens) of the model with 10 epochs. 
We set the learning rate 5e-5 for M2M, and 1e-5 for mBART-T and mBART. 
 %
\paragraph{Low-resource}
In the low-resource scenario, we use the prompt-tuning strategy only to tune the soft prompt tokens as suggested by~\citet{prefix-tuning,soft-prompt}. 
Following~\cite{ppt}, we set training epochs to 30 and learning rate to 1e-2 for all models.
\subsection{Evaluation Metrics}
\paragraph{Automatic Metrics}
\label{Automatic Metrics}
To measure the semantic quality of generated trans-lingual definitions, we apply the sentence-transformer toolkit~\cite{reimers2020making} to calculate the semantic similarity between the generated definition in the target language and the golden reference in its original language (e.g., for En-Zh, we calculate semantic similarity between generated Chinese definition and the golden English definition).

\paragraph{Manual Evaluation}
\label{Manual Evaluation} 
We also perform manual evaluation on the test set of 200 examples in low-resource setting. 
Based on the automatic evaluation results from Table~\ref{Main Results}, 
we only manually assess M2M model with three methods (Directly Fine-tuning, Prompt Combination, Prompt Contrastive Learning) in rich-resource setting, and M2M model with Prompt Contrastive Learning method in low-resource setting.

We ask six college students who achieved a score above 580 in the College English Test 6 level (CET-6) as annotators. 
Three of these students will be responsible for annotating En-Zh results, while the remaining three will focus on Zh-En results. 
Similarly, we recruit six annotators who have passed Test national du français enseigné à titre de spécialité, niveau IV (TFS-4). Three of these annotators will be assigned to annotate En-Fr and Fr-En results, the remaining three will be responsible for Zh-Fr and Fr-Zh results.

Each annotator is asked to evaluate the generated trans-lingual definitions on two aspects: (1) Accuracy (Acc.) is a measure of the semantic relevance of the definitions to the word; (2) Fluency (Flu.) evaluates their readability without considering semantic alignment. Both criteria have a range of 1-5. In addition, the annotators are asked to rate the Ignore-task error and Language-mix error.
We average the scores as the final score, and the agreements among the annotators of En-Zh, Zh-En, En-Fr \& Fr-En, and Zh-Fr \& Fr-Zh are ICC 0.937 (p<0.001), ICC 0.932 (p<0.001), ICC 0.904 (p<0.001) and 0.929 (p<0.001) respectively.

\subsection{Main Results}
\label{Main Results}
We begin by examining the automatic evaluation results in rich-resource settings. As shown in Table~\ref{Automatic-results},  
applying Contrastive Prompt Learning method on M2M and mBART-T models outperform other strategies across En-Zh and Zh-En scenarios. Furthermore, the baseline Pipeline method exhibits a performance degradation of 1.84 (En-Zh) and 2.95 (Zh-En) on the Semantic Sim metric when compared to our best method. This suggests that \textbf{the proposed Trans-lingual Definition Generation (TLDG) task cannot be simply addressed with a naive pipeline method}, which can be attributed to the errors accumulated during the pipeline.

Comparing the rows of M2M and mBART-T, M2M-based is superior on TLDG. We conjecture the superior performance comes from M2M's translation ability, which is empirically validated in~\citet{M2M}. Since M2M model is trained with massive parallel translation data and equipped with the Language-Specific Sparse technique, it is shown more powerful than mBART-T on translation tasks. The comparison between M2M and mBART-T gives us a hint that \textbf{model's translation ability has an impact on our TLDG task}, which we analyze in later sections.  

When checking the manual evaluation results in Table~\ref{manual evaluation-1}, it is notable that the proposed Contrastive Prompt Learning method obtains the highest scores on both Acc. and Flu. metrics. 
Comparing baseline Pipeline method with Contrastive Prompt Learning method in the Zh-En trans-lingual scenario (row 2 and row 8), we can see that Contrastive Prompt Learning method significantly improves trans-lingual quality, as it achieves 7.2\% relative increase on Acc and 7.1\% relative increase on Flu. A similar result in low-resource setting can refer to Appendix~\ref{Human Evaluation of Pipeline strategy in Low-resource Setting}. 
\begin{table}[!h]\small
\fontsize{9pt}{8.5pt}\selectfont
\centering
\setlength{\tabcolsep}{0.0mm}{
\begin{tabular}{@{}cccc@{}}
\toprule
Method    & \begin{tabular}[c]{@{}l@{}} Language\\ Combination\end{tabular} & \begin{tabular}[c]{@{}c@{}}Acc. $\uparrow$\end{tabular} & \begin{tabular}[c]{@{}c@{}}Flu. $\uparrow$\end{tabular}  \\ \midrule

\multirow{2}{*}{\begin{tabular}[c]{@{}c@{}}Pipeline\\(rich-resource)\end{tabular}}   & En-Zh     & 3.09 & 3.34              \\
& Zh-En    & 3.18 & 3.52      \\ \midrule

\multirow{2}{*}{\begin{tabular}[c]{@{}c@{}}w/ Directly Fine-tuning\\(rich-resource)\end{tabular}}   & En-Zh      & 3.02 & 3.37               \\
& Zh-En   & 3.08 & 3.61     \\ \midrule

\multirow{2}{*}{\begin{tabular}[c]{@{}c@{}}w/ Prompt Combination \\(rich-resource)\end{tabular}}    & En-Zh  & 3.13 & 3.45                      \\
 & Zh-En    & 3.17    & 3.67   \\ \midrule
 
\multirow{2}{*}{\begin{tabular}[c]{@{}c@{}}w/ Contrastive Prompt\\(rich-resource)\end{tabular}}                                                          & En-Zh                                                      & \textbf{3.29}$_{(+6.4\%)}$  & \textbf{3.51}$_{(+5.1\%)}$                                                                   \\
& Zh-En                                                     & \textbf{3.41}$_{(+7.2\%)}$     & \textbf{3.77}$_{(+7.1\%)}$                                             \\ \midrule
\multirow{6}{*}{\begin{tabular}[c]{@{}c@{}}w/ Contrastive Prompt\\ (low-resource)\end{tabular}} & En-Zh                                                          & 2.98 & 3.31                                                                                \\
& Zh-En         &3.08                                                 & 3.59                                                            \\
& En-Fr                    &3.04                                      & 3.48                                                         \\
& Zh-Fr         &3.07                                                 & 3.45                                                                \\
& Fr-En     &3.11                                                      & 3.62                                                            \\
& Fr-Zh       &3.02                                                   & 3.32                                                         \\ \midrule
\end{tabular}
}
\caption{Manual evaluation for quality assessment of trans-lingual definitions generated by M2M in low-resource test datasets}
\vspace{-2mm}
\label{manual evaluation-1}
\end{table}

Another interesting finding comes when we compare the performances in rich- and low-resource scenarios. Take Zh-En trans-lingual task for example. It is observed that leveraging Contrastive Prompt Learning method in low-resource setting (row 10) is comparable to the simple implementation of directly fine-tuning (row 4) in rich-resource settings. Similar findings can also be found on the rows of En-Zh trans-lingual task. These findings greatly show the potential of the proposed method in the low-resource scenario.
The results presented in Table~\ref{manual evaluation-2} demonstrate that \textbf{our Contrastive Prompt Learning method effectively mitigates the two types of errors}. Specifically, when compared to directly fine-tuning implementation in the En-Zh scenario (row 1 and row 5), the Contrastive Prompt Learning method achieves a relative decrease of 77.8\% in Language-mix error rate and perform well in Ignore-task error rate. 
\begin{table}[!h]
\fontsize{8.5pt}{12pt}\selectfont
\centering
\setlength{\tabcolsep}{0.0mm}{
\begin{tabular}{@{}cccc@{}}
\toprule
\fontsize{9pt}{12pt}\selectfont Method    & \fontsize{8.2pt}{12pt}\selectfont\begin{tabular}[c]{@{}l@{}} Language\\ Combination\end{tabular} & \fontsize{8.2pt}{12pt}\selectfont\begin{tabular}[c]{@{}c@{}}Language-mix\\ error rate$\downarrow$\end{tabular} & \fontsize{8.2pt}{12pt}\selectfont\begin{tabular}[c]{@{}c@{}}Ignore-task\\ error rate$\downarrow$\end{tabular} \\ \midrule

\multirow{2}{*}{\fontsize{8pt}{12pt}\selectfont\begin{tabular}[c]{@{}c@{}}w/ Direct Fine-tuning \\(rich-resource)\end{tabular}}   & En-Zh   & $\text{-}$       & 11.25\%              \\
& Zh-En    & $\text{-}$    & 9.50\%   \\ \midrule

\multirow{2}{*}{\fontsize{7.7pt}{12pt}\selectfont\begin{tabular}[c]{@{}c@{}}w/ Prompt Combination \\(rich-resource)\end{tabular}}    & En-Zh       & 3.50\%                                                           & 7.50\%$_{(-33.3\%)}$                                                              \\
 & Zh-En   & 4.00\% & 6.00\%$_{(-36.8\%)}$\\ \midrule
 
\multirow{2}{*}{\fontsize{8pt}{12pt}\selectfont\begin{tabular}[c]{@{}c@{}}w/ Contrastive Prompt\\(rich-resource)\end{tabular}}                                                          & En-Zh                                                           & $\text{-}$                                                        & 2.50\%$_{(-77.8\%)}$                                                              \\
& Zh-En        & $\text{-}$                                                                 & 2.00\%$_{(-78.9\%)}$                                                            \\ \midrule
\end{tabular}
}
\caption{Manual evaluation results of the two errors in trans-lingual definition generated by M2M in low-resource test datasets.}
\vspace{-5mm}
\label{manual evaluation-2}
\end{table}
\subsection{Ablation Study}
\label{Ablation Study}
\paragraph{Pooling Function}
To examine the variants of pooling functions as introduced in Section~\ref{Contrastive Prompt Learning}, we then conduct an ablation study on M2M model with the best task-ratio 0.2 obtained in Section~\ref{Analysis on Hyper-Parameter}. 

As Table~\ref{pooling-method-ablation} shows, the attention-pooling function outperforms mean- and max- pooling functions on all the metrics. The reason lies in the distinctness of how these pooling functions gather token information. When constructing task prompt representation, the attention-pooling function aggregates all the task prompt tokens with the attention weight between the task and language prompt. Intuitively, the attention weight measures the degree of language information in each token of the task prompt. 
As a result, the task prompt representation based on attention-pooling contains more precise mixed language information, and in turn aids in separating language information when implementing Prompt Contrastive Learning. 
The variations observed in different pooling functions suggest that \textbf{the approach used to obtain an accurate representation is crucial in contrastive learning}.
\begin{table}[h!]\small
\centering
\fontsize{9pt}{8.5pt}\selectfont
\setlength{\tabcolsep}{1.5mm}{
\begin{tabular}{@{}cccccccc@{}}
\toprule
\multirow{2}{*}{Model \& Method}         & \multirow{2}{*}{Pooling Function}     & \multicolumn{2}{c}{Semantic Sim} \\ \cmidrule(l){3-4}
\multirow{6}{*}{\begin{tabular}[c]{@{}c@{}}M2M\\ /w Contrastive Prompt\\ /w Task Ratio 0.2\end{tabular}} &            & En-Zh           & Zh-En          \\ \midrule
& attention     & \textbf{49.42}  & \textbf{55.19} \\
 & mean         & 48.91           & 54.75          \\
 & max         & 48.83           & 54.68          \\ \midrule
\end{tabular}}
\caption{Ablation study results on the pooling functions. The best numbers are in \textbf{bold}.}
\vspace{-5mm}
\label{pooling-method-ablation}
\end{table}


\begin{table}[h!]\small
\centering
\fontsize{9pt}{8pt}\selectfont
\setlength{\tabcolsep}{2.5mm}{
\begin{tabular}{@{}cccccccc@{}}
\toprule
\multirow{2}{*}{Model \& Method} & \multirow{2}{*}{Task Ratio} & \multicolumn{2}{c}{Semantic Sim} \\ \cmidrule(l){3-4} 
\multirow{8}{*}{\begin{tabular}[c]{@{}c@{}}M2M\\ /w Contrastive Prompt\\ /w Attention Pooling\end{tabular}} &      & En-Zh           & Zh-En          \\ \midrule 
 & 0.1        & 48.81           & 55.12          \\
 & 0.2        & \textbf{49.42}           & \textbf{55.19}          \\
& 0.3         & 48.24           & 54.76          \\
& 0.4         & 47.63           & 53.81          \\
& 0.5         & 47.87           & 53.79          \\ \midrule 
\end{tabular}}
\caption{Hyper-parameter analysis results on the task ratio. The best results are in \textbf{bold}.}
\vspace{-5mm}
\label{task-ratio}
\end{table}

\paragraph{Hyper-Parameter}
\label{Analysis on Hyper-Parameter}
Another influential factor in our method is hyper-parameter $\lambda$ in Eq.~\ref{eq:Final_Loss}. To explore its effect, 
we keep using attention-pooling in all settings and set different $\lambda$ for each model to observe the performance change.

As Table~\ref{task-ratio} shows, when the task ratio is set to 0.2, the proposed method yields the best performance. When the task ratio is lower or higher than 0.2, the performances deteriorate.
We conjecture that our model requires more generation loss to guide contrastive learning in the right way.

\subsection{Case Study}
\label{Case Study}
For better understanding, we present some cases under the rich-resource setting to vividly analyze the superiority of our Contrastive Prompt Learning method. 
Table~\ref{Error cases} compares all methods on two trans-lingual scenarios. 
After examining the definitions produced by the directly fine-tuning implementation, we find undesired words like \emph{\begin{CJK*}{UTF8}{gbsn}``经济''\end{CJK*}} (\textit{economy}) (in the En-Zh case), as well as the words \emph{``interdependence''} and \emph{``country''} (in the Zh-En case). 
All these words are the direct translations of the context words rather than the definitions.
In the Zh-En case, it is clear that the definition from the Prompt Combination method contains Language-mix error, as it includes a Chinese word \begin{CJK*}{UTF8}{gbsn}``形容''\end{CJK*} (\textit{describe}). 
In the En-Zh case, the definition produced by the baseline Pipeline method includes an unsuitable explanation word \begin{CJK*}{UTF8}{gbsn}``上升运动''\end{CJK*} (\textit{upward movement}), which might be resulted from the limited definition style's data in the translation model's training corpus.
In contrast, the Contrastive Prompt Learning method's output, which includes \begin{CJK*}{UTF8}{gbsn}``正面发展''\end{CJK*} (\textit{positive development}) and \emph{``fewer or greater''}, accurately represents the meaning of the target words. Drawing on the highest scores in Table~\ref{Automatic-results} and Table~\ref{manual evaluation-1}, we safely conclude that the proposed \textbf{Prompt Contrastive Learning is more effective in trans-lingual definition generation}.

\begin{table}[!h]
\centering
\fontsize{9pt}{9pt}\selectfont
\setlength{\tabcolsep}{1mm}{
\begin{tabular}{@{}l|l@{}}
\toprule
\emph{Word}      & upturn                                          \\\hline
\emph{Context}   & \begin{tabular}{@{}l@{}}... in response to the economic up-\\turn helped by a recovery of key\\western export markets.\end{tabular}                                                           \\\hline
\emph{Pipeline} & \begin{tabular}{@{}l@{}}\begin{CJK*}{UTF8}{gbsn}某人或某物的状况中的上升运动\end{CJK*}\\(\textit{The upward movement in the condi-}\\\textit{tion of someone or something.})\end{tabular}   
         \\\hline
\emph{Directly Fine-tuning} & \begin{tabular}{@{}l@{}}\begin{CJK*}{UTF8}{gbsn}经济的好转。\end{CJK*}\\(\textit{The improvement of the economy.})\end{tabular}   
         \\\hline
\emph{Prompt Combination} &\begin{tabular}{@{}l@{}} \begin{CJK*}{UTF8}{gbsn}形容\end{CJK*} (\textit{Describing}) a rising trend \\of something.\end{tabular}\\\hline
\emph{Contrastive Prompt} & \begin{tabular}{@{}l@{}}\begin{CJK*}{UTF8}{gbsn}比喻特定事件向正面发展。\end{CJK*}\\(\textit{The specific event is developing} \\ \textit{towards a positive direction})\end{tabular}
\\ \midrule
\midrule
\emph{Word}  & \begin{CJK*}{UTF8}{gbsn}日益\end{CJK*} (\textit{day by day})                                                                  \\\hline
\emph{Context} & \begin{tabular}{@{}l@{}}\begin{CJK*}{UTF8}{gbsn}... 各国相互依赖程度日益加深。\end{CJK*}\\ (\textit{... the degree of interdependence}\\ \textit{among countries is increasingly} \\ \textit{deepening.})\end{tabular}
\\\hline
\emph{Pipeline} & \begin{tabular}{@{}l@{}}the degree is deepening.\end{tabular}   
         \\\hline
\emph{Directly Fine-tuning} &   \begin{tabular}{@{}l@{}}increasing interdependence of\\ country.\end{tabular}
                      \\\hline
\emph{Prompt Combination} & \begin{tabular}{@{}l@{}}in a gradual and increasing degree.
\end{tabular}         \\ \hline
\emph{Contrastive Prompt} &   \begin{tabular}{@{}l@{}}to an ever greater or fewer degree.\end{tabular}
                      \\
\bottomrule
\end{tabular}}
\caption{Generated result comparison between four methods on M2M model.}
\label{Error cases}
\end{table}


\begin{table}[!h]\small
\fontsize{9pt}{9pt}\selectfont
\centering
\setlength{\tabcolsep}{1mm}{
\begin{tabular}{p{1.3cm}|p{6cm}}
\toprule
\emph{Word}      & accent                                          \\\hline
\emph{Context}   & \begin{tabular}{@{}l@{}}... cobalt blue was used to accent certain ele-\\ments including ...\end{tabular}                                                        \\\hline
\emph{M2M} & \begin{CJK}{UTF8}{gbsn}强调特定对象。\end{CJK}(\textit{Emphasize specific objects.})    \\\hline           \emph{mBART-T} & \begin{CJK}{UTF8}{gbsn}强调的重点。\end{CJK}(\textit{Key points to emphasize.})                                   \\\midrule
\midrule

\emph{Word}      &    \begin{CJK*}{UTF8}{gbsn}珍惜\end{CJK*} (\textit{cherish})                                      \\\hline
\emph{Context}   & \begin{tabular}{@{}l@{}}\begin{CJK*}{UTF8}{gbsn}..., 什么又是值得你去珍惜的？\end{CJK*}\\(\textit{..., what is worth cherishing for you?})\end{tabular}                                         \\\hline
\emph{M2M} & deeply regard the value of something.
         \\\hline
\emph{mBART-T} & regard with great appreciation.                                       \\ 
\bottomrule
\end{tabular}}
\caption{Generated result comparison between M2M based and mBART-T based models.}
\vspace{-5mm}
\label{mBART and M2M}
\end{table}              

We also conduct case studies on the choice of multi-lingual translation model, as a complementary assessment to the results in Table~\ref{Automatic-results}. 
As shown in Table~\ref{mBART and M2M}, the generated definitions of mBART-T contain \begin{CJK*}{UTF8}{gbsn}``重点''\end{CJK*} (\textit{key}) and \emph{``appreciation''}, which are not accurate for explaining the corresponding words' meanings. However, the M2M model handles these cases well. This case study further demonstrates the hint that \textbf{the translation capability of the backbone model is crucial for trans-lingual definition generation}. For more cases in both rich- and low-resource scenarios, please kindly refer to Appendix~\ref{Generated Results}.

\section{Conclusions}
In this work, we propose a novel and challenging task TLDG that generates the trans-lingual definition in an unsupervised manner. To tackle the task, we leverage multi-lingual translation models and propose an effective method of Contrastive Prompt Learning for the task. Through extensive experiments, we validate the method is capable of addressing typical errors and promising in both rich- and low-resource scenarios. In the future, we will develop more strategies to improve the quality of trans-lingual definitions.

\section*{Limitations}
\label{Limitations}
Our work has several limitations. In terms of method generalization, the proposed method depends on multi-lingual neural machine translation models to generate trans-lingual definitions, and hence limits its application scope to those languages rarely supported by translation models. Moreover, our findings are based on three languages, but different families of languages may exhibit distinct phenomenon that even challenges our conclusions.

\bibliography{anthology,custom}
\bibliographystyle{acl_natbib}

\appendix
\onecolumn
\section{Chrome Extension Application Scene}
\label{Chrome Extension Application Scene}

\begin{figure}[!h]
    \centering
    \includegraphics[scale=0.35]{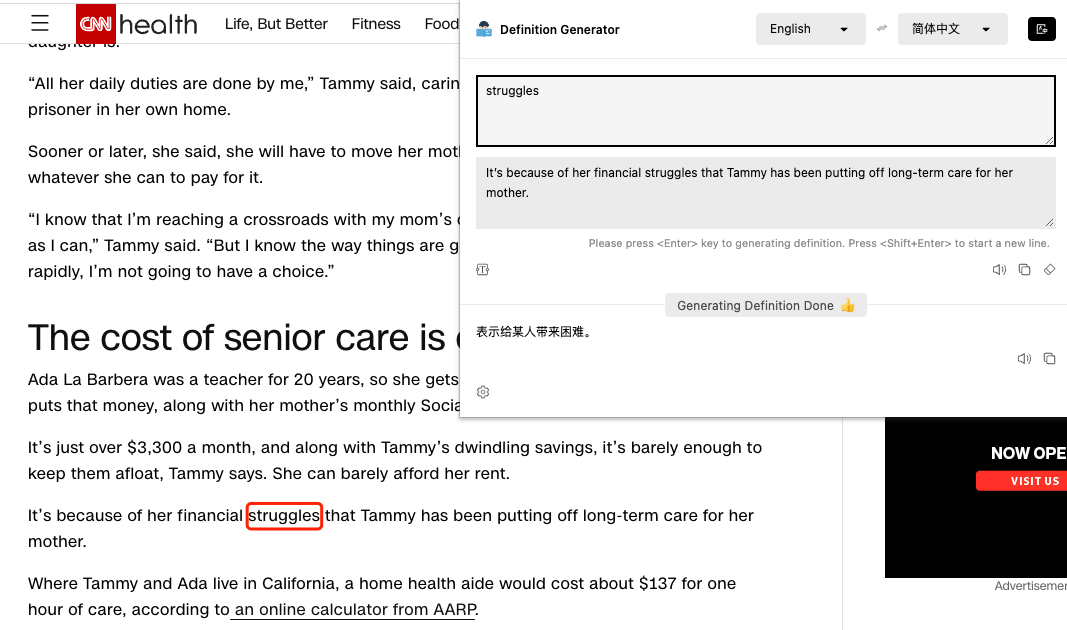}
    \caption{The application scene of Learning English words based on our best method. Given the word ``struggles'' and press the shortcut key, the application will identify its corresponding context and output the definition ``\begin{CJK*}{UTF8}{gbsn}表示给某人带来困难。\end{CJK*}'' (To make someone difficult.).}
    \label{en2zh_application}
\end{figure}

\begin{figure}[!h]
    \centering
    \includegraphics[scale=0.4]{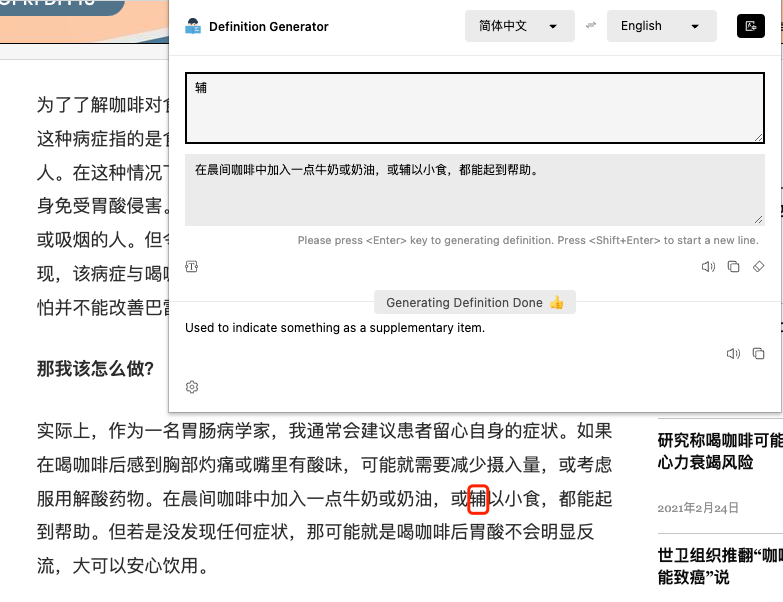}
    \caption{The application scene of Learning Chinese words based on our best method. Select the word ``\begin{CJK*}{UTF8}{gbsn}辅\end{CJK*}'' (supplement) and press the shortcut key, the application will identify its corresponding context and output the definition ``Describing something as an accessory or auxiliary item''.}
    \label{zh2en_application}
\end{figure}

\onecolumn
\section{Rich-resource Detailed Dataset Setting}
\label{Rich-resource Detailed Dataset Setting}
\begin{table*}[h!]
\centering
\begin{tabular}{lcccccc}
\hline
               & \multicolumn{3}{c}{Oxford}                                                       & \multicolumn{3}{c}{CWN}                                                          \\ \cline{2-7} 
\textbf{}      & \multicolumn{1}{l}{Train} & \multicolumn{1}{l}{Valid} & \multicolumn{1}{l}{Test} & \multicolumn{1}{l}{Train} & \multicolumn{1}{l}{Valid} & \multicolumn{1}{l}{Test} \\ \hline
Words          & 33128                     & 8,867                     & 3881                     & 6574                      & 823                       & 823                      \\
Entries        & 97,855                    & 12,232                    & 5111                     & 67861                     & 8082                      & 8599                     \\
Context length & 17.74                     & 17.80                     & 16.24                    & 34.49                     & 34.73                     & 34.04                    \\
Desc. length   & 11.02                     & 10.99                     & 10.03                    & 14.76                     & 14.60                     & 14.72                    \\ \hline
\end{tabular}
\caption{Statistics of the Oxford (English) dataset and CWN (Chinese) dataset.}
\label{dataset-statics}
\end{table*}
We use Oxford and CWN definition generation datasets in rich-resource setting experiment, the statistics of Oxford and CWN are shown in Table ~\ref{dataset-statics}.

\section{Human Evaluation of Pipeline method in Low-resource Setting}
\label{Human Evaluation of Pipeline strategy in Low-resource Setting}
\begin{table*}[h!]\normalsize
\centering
\begin{tabular}{ccccc}
\hline
\multirow{3}{*}{\begin{tabular}[c]{@{}l@{}}Language\\ Combination\end{tabular}} & \multicolumn{4}{c}{Method}                                                 \\ \cline{2-5} 
                                                                                & \multicolumn{2}{c}{\begin{tabular}{@{}l@{}} /w Contrastive\\Prompt\end{tabular}} & \multicolumn{2}{c}{Pipeline} \\ \cline{2-5} 
                                                                                & Acc                   & Flu                   & Acc           & Flu          \\ \hline
En-Zh                                                                           & 2.98                  & 3.31                  & 2.73          & 3.09         \\ 
Zh-En                                                                           & 3.08                  & 3.59                  & 2.91          & 3.34         \\ 
En-Fr                                                                           & 3.04                  & 3.48                  & 2.88          & 3.31         \\ 
Fr-En                                                                           & 3.11                  & 3.62                  & 2.98          & 3.46         \\ 
Zh-Fr                                                                           & 3.07                  & 3.45                  & 2.92          & 3.27         \\ 
Fr-Zh                                                                            & 3.02                  & 3.32                  & 2.74          & 3.13         \\ \hline
\end{tabular}
\caption{Human evaluation results of M2M /w Contrastive Prompt Learning method and baseline Pipeline method in low-resource setting.}
\label{pipeline-low-resource}
\end{table*}
We also compare our proposed M2M /w Contrastive Prompt Learning method with baseline Pipeline method in low-resource setting, the results are shown in Table~\ref{pipeline-low-resource}.

\onecolumn
\section{Generated Results}
\label{Generated Results}
\subsection{Rich-Resource Generated Results}
\begin{table*}[h!]
\centering
\begin{tabular}{@{}ll@{}}
\toprule
\emph{Word}      & telex                                          \\
\emph{Context}   & they telexed the company denying breach of contract.                                                          \\
\emph{Generated Result} & \begin{CJK*}{UTF8}{gbsn}以电传方式传送讯息。\end{CJK*}                                           \\ \midrule
\emph{Word}      & bulky    \\          
\emph{Context}   & \begin{tabular}{@{}l@{}}radio could communicate between cities, but they were too bulky to be\\ man-carried.\end{tabular}                                \\        
\emph{Generated Result} & \begin{CJK*}{UTF8}{gbsn}形容体积大的。\end{CJK*}                                  \\ \midrule

\emph{Word}      & concession                                          \\
\emph{Context}   & a corona and one adverb of resignation - or is it concession?                                                           \\
\emph{Generated Result} & \begin{CJK*}{UTF8}{gbsn}承认或授权后述对象。\end{CJK*}                                           \\ \midrule
\emph{Word}      & electronic                                          \\
\emph{Context}   & 1987 was an early but fertile time for electronic dance music.                                                           \\
\emph{Generated Result} & \begin{CJK*}{UTF8}{gbsn}以电子方式进行演奏。\end{CJK*}                                    \\ \midrule                                
\emph{Word}      & spiral                                          \\
\emph{Context}   & tensions have spiraled between pyongyang and the us.                                                           \\
\emph{Generated Result} & \begin{CJK*}{UTF8}{gbsn}比喻特定事件在一段长时间内持续进行。\end{CJK*} 
           \\ \midrule
 \emph{Word}      & fortune                                          \\
\emph{Context}   & I have had the good fortune to see the piece several times.                                \\
\emph{Generated Result} & \begin{CJK*}{UTF8}{gbsn}形容运气好。\end{CJK*}    
\\ \midrule
\emph{Word}      & revitalize                               \\        
\emph{Context}   & this food revitalized the patient.                                  \\
\emph{Prompt Combination} & \begin{CJK}{UTF8}{gbsn}使后述对象恢复生命力。\end{CJK}                                                                                \\ 
\midrule
\emph{Word}  & \begin{CJK}{UTF8}{gbsn}意外\end{CJK}                                          \\                        
\emph{Context} & \begin{tabular}{@{}l@{}}\begin{CJK}{UTF8}{gbsn}好在我们都已买了保险，如果发\end{CJK}\begin{CJK}{UTF8}{gbsn}生意外，一切都由保险公司理赔。\end{CJK}\end{tabular}
\\
\emph{Generated Result} & \begin{tabular}{@{}l@{}}an unfortunate or unexpected occurrence of something.\end{tabular}     
                           \\ \midrule
\emph{Word}      & \begin{CJK*}{UTF8}{gbsn}学术\end{CJK*}                                         \\
\emph{Context}   & \begin{tabular}{@{}l@{}}\begin{CJK*}{UTF8}{gbsn}国立大学及所有私校没必要一窝蜂搞学术，现在学生所学和社会\end{CJK*}\\ \begin{CJK*}{UTF8}{gbsn}往往都是脱节的。\end{CJK*}\end{tabular}                                \\
\emph{Generated Result} & an academic activity of the university or community.                                               \\ \midrule
\emph{Word}      & \begin{CJK*}{UTF8}{gbsn}立国\end{CJK*}                                         \\
\emph{Context}   & \begin{CJK*}{UTF8}{gbsn}立国精神、民族意识的观念如果不在军训课中提醒学生，根本没有机会。\end{CJK*}                                \\
\emph{Generated Result} & the establishment of state.                                               \\ \midrule
\emph{Word}      & \begin{CJK*}{UTF8}{gbsn}近\end{CJK*}                                         \\
\emph{Context}   & \begin{CJK*}{UTF8}{gbsn}包公庙..., 近因电视剧包青天脍炙人口而引来参拜人潮。\end{CJK*}                                \\
\emph{Generated Result} & to indicate that not a long time ago.                                    \\ \midrule
\emph{Word}      & \begin{CJK*}{UTF8}{gbsn}维\end{CJK*}                                         \\
\emph{Context}   & \begin{tabular}{@{}l@{}}\begin{CJK*}{UTF8}{gbsn}怪手及人员到市场附近巡视，凡发现摊架，则一律予以铲除，以\end{CJK*}\\ \begin{CJK*}{UTF8}{gbsn}维公权力的威信。\end{CJK*} \end{tabular}                               \\
\emph{Generated Result} & maintain the state of ( something ).                                               \\
\bottomrule
\end{tabular}
\caption{The generated results of M2M model with Contrastive Prompt Learning method under rich-resource setting.}
\label{Rich-Results}
\end{table*}

\onecolumn
\subsection{Low-Resource Generated Results}
\begin{table*}[!h]
\centering
\begin{tabular}{@{}ll@{}}
\toprule
\emph{Word}      & concession                                          \\
\emph{Context}   & a corona and one adverb of resignation - or is it concession?                                                           \\
\emph{Generated Result} & \begin{CJK*}{UTF8}{gbsn}形容被授权的。\end{CJK*} 
\\ \midrule
\emph{Word}      & antithesis                                          \\
\emph{Context}   & his theory is the antithesis of mine.                                                          \\
\emph{Generated Result} & \begin{CJK*}{UTF8}{gbsn}形容与特定事件相反的。\end{CJK*} 
\\ \midrule
\emph{Word}      & conditional                                          \\
\emph{Context}   & the conditional sale will not be complete until the full purchase price is paid.                                                        \\
\emph{Generated Result} & \begin{CJK*}{UTF8}{gbsn}形容有条件的。\end{CJK*} 
\\ \midrule
\emph{Word}      & lame                                          \\
\emph{Context}   & the comedy aspect is a little lame, with too many one-liners                                                           \\
\emph{Generated Result} & \begin{CJK*}{UTF8}{gbsn}形容缺乏活力的。\end{CJK*}                                           \\ \midrule
\emph{Word}      & surge                                          \\
\emph{Context}   & \begin{tabular}{@{}l@{}}the testing equipment-maker 's shares surged as sales rose for the first \\time in six quarters.\end{tabular}                                                           \\
\emph{Generated Result} & \begin{CJK*}{UTF8}{gbsn}形容特定对象数量增加。\end{CJK*} 
           \\ \midrule
\emph{Word}      & \begin{CJK*}{UTF8}{gbsn}近\end{CJK*}                                         \\
\emph{Context}   & \begin{tabular}{@{}l@{}}\begin{CJK*}{UTF8}{gbsn}我认为太阳在清早刚出来的时候离人近，中午的时候离人远。\end{CJK*}\end{tabular}                               \\
\emph{Generated Result} & close to or nearby.                                         \\ \midrule
\emph{Word}      & \begin{CJK*}{UTF8}{gbsn}意外\end{CJK*}                                         \\
\emph{Context}   & \begin{CJK*}{UTF8}{gbsn}好在我们都已买了保险，如果发生意外，一切都由保险公司理赔。\end{CJK*}                                \\
\emph{Generated Result} & an accidental occurrence.                                               \\ \midrule
\emph{Word}      & \begin{CJK*}{UTF8}{gbsn}看法\end{CJK*}                                         \\
\emph{Context}   & \begin{CJK*}{UTF8}{gbsn}我希望七月初开院士会议时，能够再提出在这方面一些具体的看法。\end{CJK*}                                \\
\emph{Generated Result} & the opinion of a person.   \\ \midrule

\emph{Word}      & \begin{CJK*}{UTF8}{gbsn}终究\end{CJK*}                                         \\
\emph{Context}   & \begin{CJK*}{UTF8}{gbsn}走在错误的路上，终究是要输的。\end{CJK*}                                \\
\emph{Generated Result} & Décrivez le résultat final de l'événement.                                              \\ \midrule
\emph{Word}      & \begin{CJK*}{UTF8}{gbsn}日益\end{CJK*}                                         \\
\emph{Context}   & \begin{CJK*}{UTF8}{gbsn}融入实际生活的经验中，人生经验便日益丰富。\end{CJK*}                                \\
\emph{Generated Result} & Le degré de description est approfondi.                                              \\ \midrule
\emph{Word}      & revitalize                                         \\
\emph{Context}   & this food revitalized the patient.                              \\
\emph{Generated Result} & Donner une nouvelle vitalité.                                              \\ \midrule
\emph{Word}      & enter                                         \\
\emph{Context}   & enter a drug treatment program.                         \\
\emph{Generated Result} & Participer à un programme ou un projet.                                             
\\
\bottomrule
\end{tabular}
\caption{The generated results of M2M model with Contrastive Prompt Learning method under low-resource setting.}
\label{Rich-Results}
\end{table*}

\end{document}